\documentclass[letterpaper,10pt,conference]{ieeeconf}
\IEEEoverridecommandlockouts
\usepackage{graphicx}
\usepackage{caption}
\usepackage{cuted}
\usepackage{amsmath,amssymb}
\usepackage{times}
\usepackage[noend]{algorithmic}
\usepackage{algorithm}
\usepackage{booktabs}
\usepackage{multirow}
\usepackage{url}
\usepackage{bbm}
\usepackage{xcolor}
\usepackage{soul}
\definecolor{lightblue}{RGB}{173,216,230}
\sethlcolor{lightblue}

\DeclareMathOperator{\Log}{Log}

\begin{document}
\title{\LARGE\bf
UniBiDex: A Unified Teleoperation \\Framework for Robotic Bimanual Dexterous Manipulation}
\author{
Zhongxuan Li$^{*}$,
Zeliang Guo$^{*}$,
Jun Hu,
David Navarro-Alarcon,
Jia Pan, 
Hongmin Wu$^{\dag}$
and Peng Zhou$^{\dag}$
\thanks{
$*$ denotes equal contribution and $\dag$ denotes corresponding author.
}
\thanks{
This work was supported by the National Natural Science Foundation of China (NSFC) under Grant No. 62403211, and in part by
National Key Research and Development Program of China (2024YFB4709700), Guangdong Basic and Applied Basic Research Foundation (2025A1515011114, 2024A1515240007).
\emph{Corresponding author: Hongmin Wu, Peng Zhou.} 
}
\thanks{
Z. Li and J. Pan are with the Department of Computer Science, The University of Hong Kong, Pokfulam, Hong Kong.
}
\thanks{
Z. Guo, J. Hu and P. Zhou are with the Dongguan Key Laboratory of Intelligent Equipment and Smart Industry, School of Advanced Engineering, The Great Bay University, Dongguan, China.
}
\thanks{
D. Navarro-Alarcon is with the Department of Mechanical Engineering, The Hong Kong Polytechnic University, Kowloon, Hong Kong.
}
\thanks{
H. Wu is with the Guangdong Academy of Sciences, Guangzhou, China.
}
}
\maketitle

{%
  \setlength\stripsep{-5.0em}
  \begin{strip}
    \centering
      \includegraphics[width=0.99\textwidth]{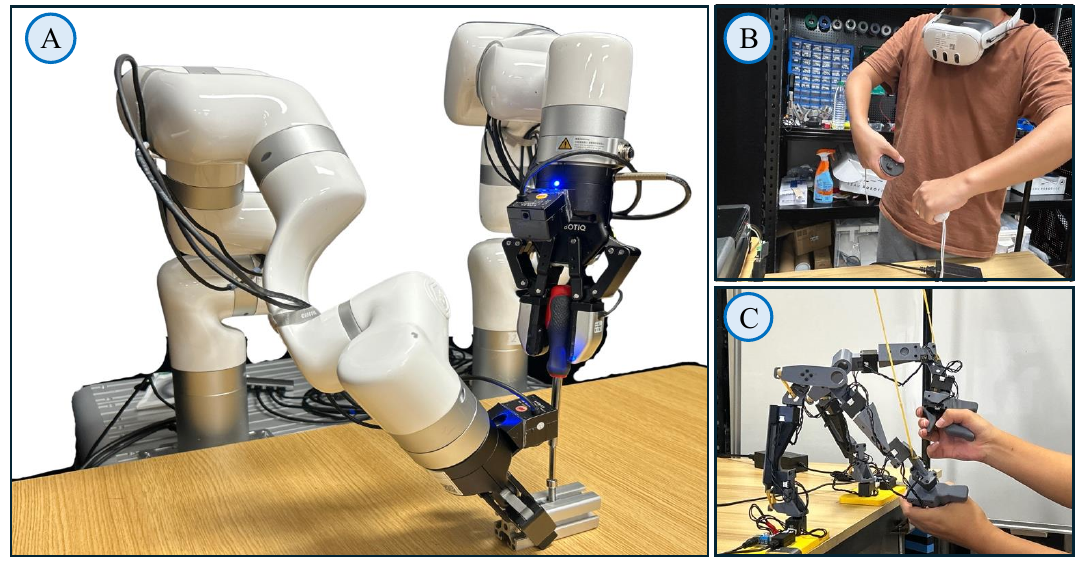}
      \captionof{figure}{We show the UniBiDex framework built for contact-rich bimanual teleoperation. The user can teleoperate the robot arms by controlling either the VR headset or leader arms. The system processes heterogeneous input devices with a unified kinematics module.}
      \label{fig:teaser}
    \vspace{6.0em}
  \end{strip}
}

\begin{abstract}
We present \textbf{UniBiDex}, a unified teleoperation framework for robotic bimanual dexterous manipulation that supports both VR-based and leader–follower input modalities. UniBiDex enables real-time, contact-rich dual-arm teleoperation by integrating heterogeneous input devices into a shared control stack with consistent kinematic treatment and safety guarantees. The framework employs null-space control to optimize bimanual configurations, ensuring smooth, collision-free and singularity-aware motion across tasks. We validate UniBiDex on a long-horizon kitchen-tidying task involving five sequential manipulation subtasks, demonstrating higher task success rates, smoother trajectories, and improved robustness compared to strong baselines. By releasing all hardware and software components as open-source, we aim to lower the barrier to collecting large-scale, high-quality human demonstration datasets and accelerate progress in robot learning.

\end{abstract}

\section{INTRODUCTION}
Robot learning methods have transformed robotics by replacing hand-crafted control policies with behaviors learned directly from data across perception, planning, and control \cite{zitkovich2023rt, black2024pi0, weng2024interactive}. Within this paradigm, imitation learning \cite{schaal1999imitation, zhang2018deep, zhou2024imitating} enables robots to acquire complex manipulation skills from human demonstrations, and its effectiveness increases with richer datasets. Recent studies \cite{mandlekar2021matters, lin2024data, chi2024universal} show that expanding datasets with diverse object types~\cite{qi2025coarse,zhang2025joint}, force interactions, and bimanual coordination patterns leads to more accurate task execution and improved generalization \cite{zhou2024bimanual}. Therefore, developing low-cost, data‑efficient teleoperation systems is critical for collecting data to train more advanced imitation learning policies. However, existing teleoperation systems often fall short of capturing the nuance and dexterity required for contact‑rich bimanual manipulation. They frequently lack coordinated dual-arm control and robust safety mechanisms, limiting the precision, speed, and overall quality of the demonstration data. Additionally, existing teleoperation systems often lack force feedback, preventing users from perceiving subtle contact forces experienced by the follower arm. Finally, many teleoperation algorithms are tied to a single modality, which restricts their extensibility across different input devices.

To overcome these limitations, we introduce \textbf{UniBiDex}: a \underline{Uni}fied Teleoperation Framework for Robotic \underline{Bi}manual \underline{Dex}terous Manipulation. Our framework enables precise, coordinated two‑arm teleoperation with built‑in safety guarantees, facilitating the scalable collection of high‑fidelity human demonstrations in contact‑rich tasks. The name “UniBiDex” reflects two core principles. First, the system offers universal device support: any input modality—such as VR controllers or leader arms—can be integrated into the same teleoperation framework, enabling user-friendly teleoperation and haptic feedback. Second, the system provides unified constraint handling: all safety and task‑related constraints are enforced through a single bimanual control module, regardless of the input modality. This module leverages null‑space control to exploit the redundancy of dual 7‑DoF arms, enabling the computation of optimal, coordination‑aware bimanual configurations. Our main contributions are summarized as follows:

\begin{itemize}
\item We present a practical implementation of UniBiDex, featuring modular software, dual‑mode hardware integration (VR and leader–follower), and a unified kinematics module for real‑time, contact‑rich bimanual teleoperation.
\item We conduct a comprehensive user study that demonstrates the system’s effectiveness in collecting complex manipulation data compared with other widely used, low‑cost teleoperation systems reported in the literature.
\item We release all hardware designs and software components as open‑source resources to ensure reproducibility and enable the research community to build upon our work.
\end{itemize}

\section{RELATED WORK}
\subsection{Teleoperation Devices}
Beyond specialized options such as 3D mice or generic 6‑DoF motion‑tracking devices \cite{darvish2023teleoperation}, the two most accessible teleoperation interfaces are VR headsets \cite{BunnyVisionPro, iyer2024open, wang2024dexcap} and leader–follower arms \cite{zhao2023learning, fu2024mobile, wu2024gello, liu2025factr, zhou2021lasesom}. VR systems are lightweight, portable, and provide an immersive first‑person view with multi‑DoF tracking—making them well‑suited for complex, free‑form tasks. However, discrepancies between human and robot kinematics can yield invalid inverse‑kinematics solutions, resulting in unintended commands that exceed joint limits, reduce demonstration throughput, and increase failures near singularities or self‑collisions. Leader–follower control employs secondary robot arms that are isomorphic to the target manipulator, enabling true one‑to‑one motion mapping and force feedback. This matched morphology allows operators to intuitively perceive and respect the robot’s kinematic constraints, enabling precise, low‑latency control. Nonetheless, these systems have notable drawbacks, including bulky hardware, restricted workspaces, and substantial integration effort, along with limited flexibility to adapt to new robots or tasks without redesign.

\subsection{Kinematics Constraints in Bimanual Manipulation}
Kinematic constraints present a significant challenge in bimanual teleoperation, and recent systems have incorporated solutions directly into control algorithms. Some approaches formulate inverse kinematics (IK) as constrained optimization problems that address singularity and collision avoidance. For example, \cite{BunnyVisionPro} introduced a VR-based dual-arm system that optimizes singularity and self-collision penalties. \cite{grannen2023stabilize} proposed a coordination heuristic that mitigates IK conflicts by assigning different roles to each arm. \cite{he2024omnih2o} presented a neural-based whole-body controller that integrates VR controller inputs to safely coordinate multiple limbs. However, key limitations remain. First, many systems are overly conservative, enforcing large safety margins that shrink the effective workspace and restrict dexterity. Second, constraint parameters are often robot- or task-specific, requiring manual tuning and limiting adaptability. Third, inter-arm coordination typically relies on simple heuristics rather than unified, real-time optimization, which can fail during tightly synchronized, contact-rich manipulation. Addressing these challenges is essential for enabling robust and adaptable bimanual teleoperation.

\section{The Methodology}
\begin{figure*}[htbp]
    \centering
    \includegraphics[width=0.99\linewidth]{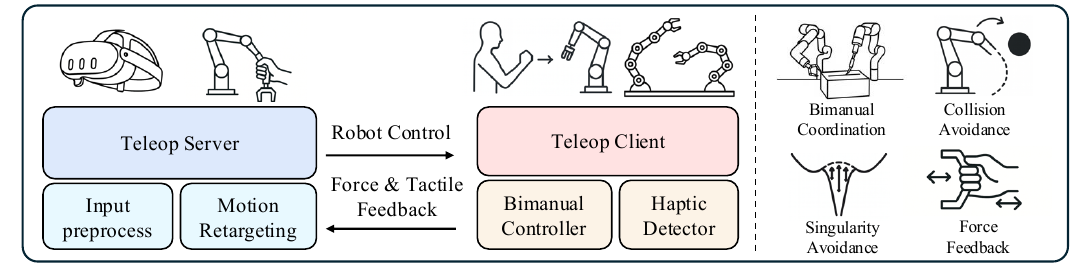}
    \caption{Illustration of the proposed teleoperation framework. The system processes input commands from heterogeneous devices and formulates task-related and safety constraints for bimanual control as a single optimization problem.}
    \label{fig:framework}
\end{figure*}

\subsection{Overall System Design}
The proposed system supports two teleoperation input modalities: VR headsets and leader–follower arms. For the VR mode, we use the Meta Quest 3 \cite{metaquest3}, while the leader–follower mode is based on the hardware design of the GELLO system \cite{wu2024gello}. A unified dual-arm control layer minimizes collisions and inverse-kinematics failures in both modes, enabling precise and reliable bimanual manipulation.

\begin{figure}[h]
\centering
\includegraphics[width=0.99\linewidth]{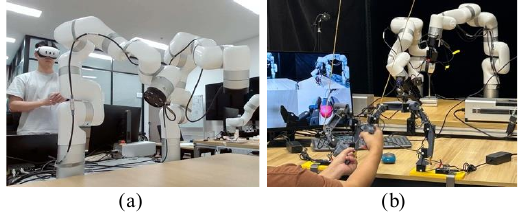}
\caption{(a) VR mode. (b) Leader–follower arm mode. UniBiDex supports both VR- and leader arm-based control with nearly identical performance.}
\label{fig:enter-label}
\vspace{-0.5em}
\end{figure}

Figure~\ref{fig:framework} illustrates the overall architecture of the proposed teleoperation framework, which consists of four decoupled modules: \textit{input preprocessing}, \textit{motion retargeting}, \textit{bimanual motion control}, and \textit{haptic feedback}. The \textit{input preprocessing} module receives the target end–effector poses \(p_{\text{in}}\) and converts them into the robot base frame. If the user operates via a leader arm, it also receives joint angles \(q_{0:k}\), where \(k\) is the number of joints. This module filters raw input to smooth spikes and suppress abnormal values, ensuring stable downstream control. The \textit{motion retargeting} module translates input poses into joint configurations via inverse kinematics. In VR-based teleoperation, this step is necessary because only end–effector poses are available. For leader–follower arms, although joint angles are directly observed, inverse kinematics is still applied to identify and correct potentially invalid configurations (e.g., self-collisions or singularities). The \textit{bimanual motion control} module computes the final joint commands by optimizing the configurations of both arms. We employ null-space control to exploit the redundancy of dual 7‑DoF arms, enabling secondary objectives such as maintaining safe inter-arm distances and maximizing manipulability. Finally, the \textit{haptic feedback} module estimates external contact forces using joint current measurements from the robot arms. After compensating for gravity effects, the processed signals are transmitted back to the teleoperation device, providing intuitive force feedback to the human operator.

\subsection{Input Preprocessing and Motion Retargeting}
To integrate heterogeneous input devices into a unified control framework, we define a \emph{virtual base frame} at the user’s initial input pose (either the VR‑controller origin or the leader‑arm base). All subsequent teleoperation commands are expressed relative to this frame and then retargeted to the robot, ensuring consistent behavior across devices and sessions.

\subsubsection{Notation}
For each side \(i\in\{L,R\}\):
\begin{itemize}
  \item \({}^{V}T_{C_i}(t)\in SE(3)\): current control pose in the virtual base frame.
  \item \({}^{V}T_{C_i}^0\): initial controller pose at \(t=0\).
  \item \({}^{R}T_{E_i}^{\mathrm{des}}(t)\in SE(3)\): desired end–effector pose.
  \item \({}^{R}T_{E_i}^0\): robot end–effector pose at \(t=0\).
\end{itemize}

\subsubsection{Relative Motion Retargeting}
We first compute the user’s relative controller motion,
\begin{equation}
  {}^{V}\Delta T_{C_i}(t)
  = ({}^{V}T_{C_i}^0)^{-1}\,{}^{V}T_{C_i}(t).
\end{equation}
The desired robot end–effector poses are then calculated as

\begin{equation}
    {}^{R}T_{E_i}^{\mathrm{des}}(t) = {}^{V}\Delta T_{C_i}(t) \, {}^{R}T_{E_i}^0
\end{equation}

\subsection{Bimanual Teleoperation Coordination}
\begin{figure}[htbp]
    \centering
    \includegraphics[width=0.99\linewidth]{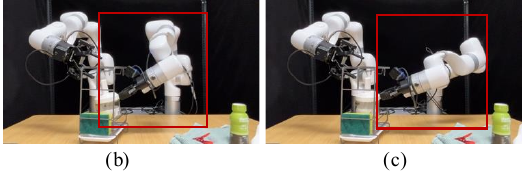}
    \caption{Illustration of the optimal reference configuration. The right arm joint configuration in (c) provides greater degrees of freedom compared to the arm in (b), resulting in more compliant control in subsequent motions.}
    \label{fig:reference-configurations}
\end{figure}

Before initiating the inverse kinematics computation, we define a set of optimal bimanual configurations, denoted as \(\mathcal{Q}_{\rm ref} = \{(q_L^{(1)}, q_R^{(1)}), \dots, (q_L^{(M)}, q_R^{(M)})\}\). These reference poses can be obtained either through dragging the follower arms or by recording motions from the leader arms. As illustrated in Figure~\ref{fig:reference-configurations}b, \(\mathcal{Q}_{\rm ref}\) captures preferred dual-arm configurations for tasks such as inserting a seasoning box into a shelf. Notably, the configuration demonstrates superior manipulability in cluttered environments, offering greater reach and collision avoidance capability.

After obtaining the desired end–effector targets for each arm through relative motion retargeting, the system computes the joint commands by solving a safety‑aware inverse kinematics (IK) optimization. For each arm $i$, at every control step we solve:

\begin{equation}
\begin{aligned}
\Delta\mathbf{q}_i^{task} = \arg\min_{\Delta \mathbf{q}} \Big(
  &\underbrace{\|J_i \Delta \mathbf{q} - \mathbf{e}_i^{\mathrm{cart}}\|^2}_{\text{(1) Cartesian tracking}} +
  \underbrace{\omega_q \|\Delta \mathbf{q} - \Delta \mathbf{q}_{C_i}\|^2}_{\text{(2) Joint matching}} \\
  &+ \underbrace{\mu^2 \|\Delta \mathbf{q}\|^2}_{\text{(3) Damping}} \Big) \\
\end{aligned}
\label{eq:joint_inc_main}
\end{equation}

where $\mathbf{e}_i^{\mathrm{cart}}=\Log({}^{R}T_{E_i}(t)^{-1}\,{}^{R}T_{E_i}^{\mathrm{des}}(t))^\vee$ represents the small incremental end‑effector motion at time $t$, and $\Delta\mathbf{q}_{C_i}$ encodes leader–follower joint increments (set to zero in VR mode),  $J_i$ denotes the geometric Jacobian of arm~$i$.

The optimization is implemented in two stages. First, an unconstrained update is performed by solving the regularized least‑squares problem in Eq.~\eqref{eq:joint_inc_main} without considering collisions, yielding a candidate increment \(\Delta\mathbf{q}_i\) for each arm \(i \in \{L, R\}\). To further improve bimanual consistency, a null‑space control term is incorporated to attract each arm toward a set of predefined optimal bimanual poses \(\mathcal{Q}_{\rm ref}\). At each control step, we select the closest full configuration pair \((q_L^*, q_R^*)\in\mathcal{Q}_{\rm ref}\) to the current robot state \((q_L, q_R)\) via:

$$
  (q_L^*, q_R^*) = \mathop{\arg\min}_{\scriptstyle (q_L^{(k)}, q_R^{(k)}) \in \mathcal{Q}_{\rm ref}} 
  \|q_L^{(k)} - q_L\|^2 + \|q_R^{(k)} - q_R\|^2.
$$

For each arm $i$, we then compute the optimal null‑space increment:

$$
\Delta q_{i,\mathrm{null}}^{opt} = \arg \min_{\Delta q_{i,\mathrm{null}}} \| q_i + \Delta q_{i,\mathrm{null}} - q_i^* \|
$$

where the null‑space increment is represented by 
$$
\Delta q_{i,\mathrm{null}} = k_n (I - J_i^\dagger J_i) \Delta q_i^{task}
$$. 

Here $J_i^\dagger = (J_i^\top J_i)^{-1} J_i^\top$ is the pseudoinverse of the arm Jacobian and \(k_n > 0\) is a scalar gain. Finally, we augment the unconstrained solution:

$$
\Delta q_i \leftarrow \Delta q_i^{task} + \Delta q_{i,\mathrm{null}}^{opt}
$$

By doing the above steps, each arm is subtly guided toward its corresponding reference pose, improving configuration consistency and reducing anomalous IK behavior without disrupting primary end–effector tracking. This unified procedure ensures accurate, real-time, and collision-aware bimanual teleoperation across all input modalities.

In practice, $Q_{\text{ref}}$ is obtained in a one-time calibration lasting 5–10 minutes per manipulation task. An expert operator uses the leader arm to teleoperate the robot and records 8–12 canonical bimanual configurations (e.g., reaching, approaching, handover) as joint-angle pairs on the followers. The resulting set is stored as a configuration file. In our experiments, we consistently used 10 reference poses without per-subtask tuning.

We did not enforce full‑body collision detection across all links of the dual‑arm robot, as doing so was found to hinder precise control in contact‑rich manipulation tasks. Instead, we observed that null‑space control naturally guided the system toward collision‑free configurations by exploiting kinematic redundancy. To maintain safety without compromising dexterity, we implemented a lightweight watchdog thread that monitors abrupt joint movements. 

Our framework ensures accurate end–effector tracking and compliant motion across all input modalities. After computing the optimal joint increments, a PD controller is applied to generate joint velocity commands, which are then transmitted to the follower arm.

\subsection{Haptic Feedback}
In the teleoperation loop, both kinesthetic and vibrotactile feedback are generated for the operator using joint current measurements from the follower arm. At each control step, motor currents from all joints are recorded. These raw signals include contributions from gravity compensation as well as external interaction forces. By subtracting the precomputed gravity torques—derived from the robot’s mass and geometry—the system isolates the net interaction torque caused by contact or payload.

This torque is rendered directly on the leader arm, allowing the operator to intuitively perceive the forces experienced by the follower arm and providing kinesthetic feedback. In parallel, the magnitude of the interaction torque is mapped to a vibration signal on the VR controllers. This vibrotactile cue complements the kinesthetic channel by highlighting subtle contact events or dynamic interactions that might otherwise go unnoticed.

We note that while recent work such as~\cite{liu2025factr} provides force feedback to the leader arm, it relies on high-end robotic platforms equipped with joint torque sensors (e.g., Franka~\cite{frankaemika}). In contrast, our approach infers external interaction forces directly from joint motor currents, enabling a more general and cost-effective solution applicable to low-cost robotic arms without built-in force sensing. This makes our system broadly accessible while still supporting rich haptic feedback.

Together, these two feedback channels create a rich, multimodal haptic interface that enhances operator awareness and improves manipulation performance in contact‑rich tasks.

\section{Experiments}
\subsection{User Study Procedure}
We recruited three male and one female participants, all with prior teleoperation experience. Each participant was asked to complete a single, long-horizon bimanual manipulation task under two input modalities: VR-based and leader–follower. The task simulates a household kitchen-tidying routine and involves sequential, contact-rich subtasks requiring dexterous coordination:

\begin{enumerate}
\item \textit{Item Unpacking:} Pick up a group of assorted items (seasoning box, drink bottle, cleaning sponge, towel, and clamp) from a cloth bag and place them on the table.
\item \textit{Shelf Organization:} Sort and place the items into their designated positions in a multi-level kitchen shelf.
\item \textit{Towel Folding:} Fold the towel neatly along a marked midline using both grippers.
\item \textit{Towel Placement:} Place the folded towel onto the rack section of the shelf.
\item \textit{Clamp Attachment:} Attach the clamp onto the rack such that it secures the towel in place.
\end{enumerate}

\begin{figure*}[h]
  \centering
  \includegraphics[width=0.99\linewidth]{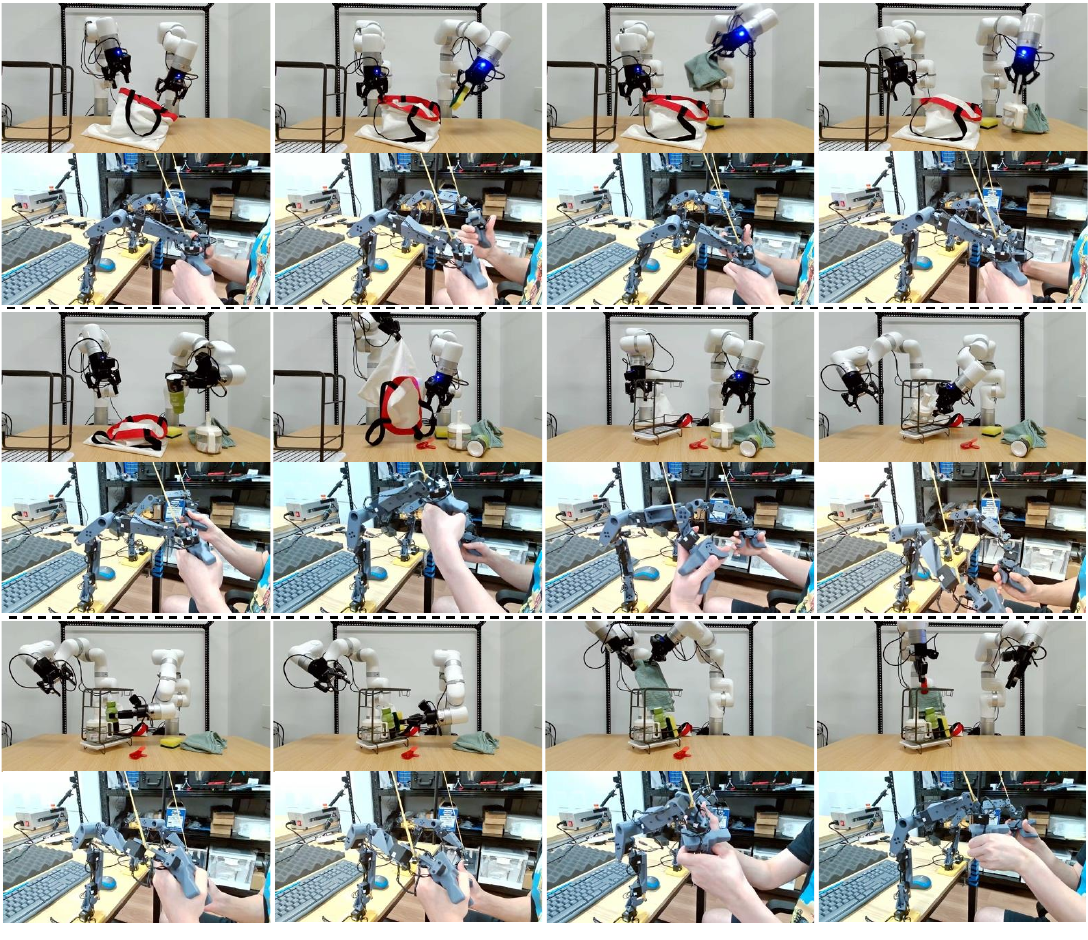}
  \caption{Overview of the leader–follower arm teleoperation workflow for the household kitchen–tidying task, showing the five sequential subtasks: (1) item unpacking, (2) shelf organization, (3) towel folding, (4) towel placement, and (5) clamp attachment.}
  \label{fig:experiment-lf}
\end{figure*}

Before starting the task, participants received a five-minute training session per modality to familiarize themselves with the control interface and virtual base frame calibration. Rest breaks were allowed between trials to prevent fatigue. All experiments were conducted on a dual-arm XArm-7 robot equipped with parallel-jaw grippers. In VR mode, participants used a Meta Quest 3 headset. We reimplemented the teleoperation logic of Bunny Vision Pro~\cite{BunnyVisionPro} and GELLO~\cite{wu2024gello} on our hardware to enable a fair and meaningful comparison. For clarity, we define the baseline implementations as follows. \emph{Naive VR}: a direct position-based IK solver mapping the VR controller pose to the robot, without null-space coupling or attraction toward $Q_\text{ref}$. \emph{Naive LF}: a direct 1:1 joint mapping from leader to follower arms with light smoothing, again without null-space coordination or $Q_\text{ref}$. In contrast, \emph{UniBiDex} employs a safety-aware IK solver with null-space attraction toward $Q_\text{ref}$, which regularizes the arms into consistent, safe bimanual postures. System performance was evaluated in terms of success rate and task completion time among all successful trials.

\subsection{Teleoperation Results}
We collected 40 trials per modality (1 long‑horizon task $\times$ 4 users $\times$ 2 modes $\times$ 5 repetitions), for a total of 80 trials. Table~\ref{tab:results_combined} reports the mean and standard deviation of completion time and success rate for each subtask. 

\begin{figure*}[htbp]
  \centering
  \includegraphics[width=0.99\linewidth]{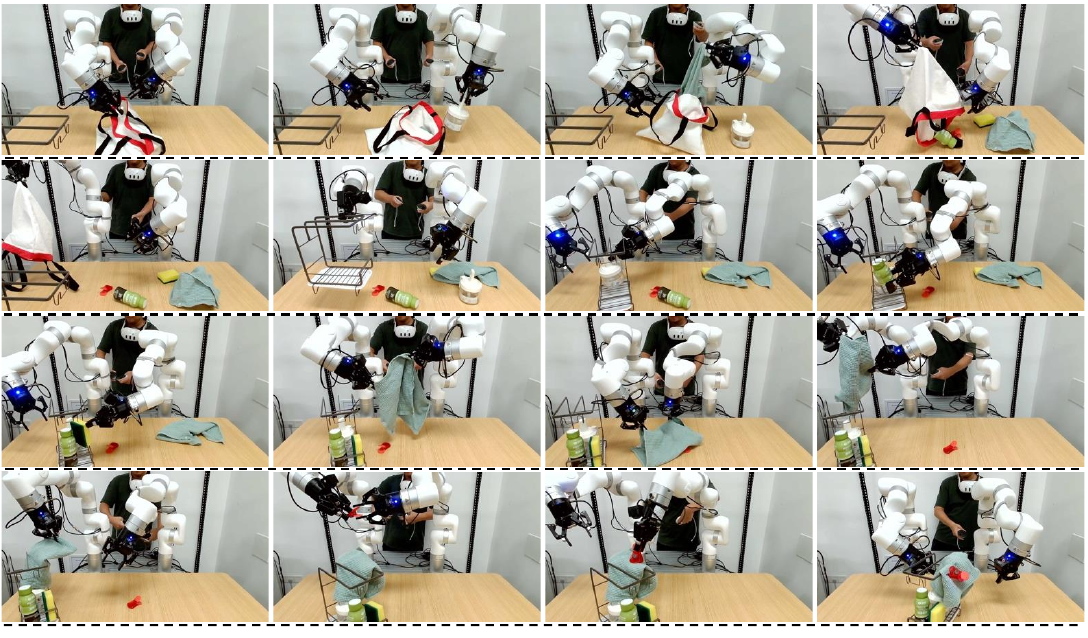}
  \caption{Overview of the VR-based teleoperation workflow for the household kitchen–tidying task.}
  \label{fig:experiment-vr}
\end{figure*}

\begin{table*}[htbp]
  \centering
  \footnotesize
  \caption{Completion time (s) and success rate across five subtasks and overall task performance}
  \label{tab:results_combined}
  \begin{tabular}{l cc cc cc cc cc cc}
    \toprule
    \multirow{2}{*}{\textbf{Method}}
      & \multicolumn{2}{c}{\textbf{Item Unpacking}}
      & \multicolumn{2}{c}{\textbf{Shelf Organization}}
      & \multicolumn{2}{c}{\textbf{Towel Folding}}
      & \multicolumn{2}{c}{\textbf{Towel Placement}}
      & \multicolumn{2}{c}{\textbf{Clamp Attachment}}
      & \multicolumn{2}{c}{\textbf{Overall Task}} \\
    & Time (s) & Succ & Time (s) & Succ & Time (s) & Succ & Time (s) & Succ & Time (s) & Succ & Time (s) & Succ \\
    \midrule
    \multicolumn{13}{c}{\emph{VR-based Methods}} \\
    \emph{UniBiDex (VR)}
      & $238\pm12$  & 31/40
      & $198\pm10$  & 30/40
      & $118\pm6$   & 32/40
      & $80\pm4$    & 24/40
      & $38\pm4$    & 24/40
      & $672\pm20$  & 24/40 \\
    \emph{Naive VR}
      & $284\pm16$  & 28/40
      & $236\pm12$  & 26/40
      & $142\pm10$  & 25/40
      & $96\pm6$    & 22/40
      & $58\pm4$    & 22/40
      & $816\pm24$  & 18/40 \\
    \midrule
    \multicolumn{13}{c}{\emph{Leader–Follower Methods}} \\
    \emph{UniBiDex (LF)}
      & $113\pm5$   & 37/40
      & $94\pm4$    & 36/40
      & $56\pm3$    & 36/40
      & $38\pm2$    & 30/40
      & $18\pm2$    & 30/40
      & $319\pm8$   & 30/40 \\
    \emph{Naive LF}
      & $117\pm6$   & 33/40
      & $97\pm5$    & 32/40
      & $57\pm4$    & 31/40
      & $42\pm3$    & 25/40
      & $22\pm2$    & 25/40
      & $335\pm9$   & 24/40 \\
    \bottomrule
  \end{tabular}
  \vspace{-0.1em}
\end{table*}

Overall, our unified framework delivered markedly more consistent performance across the entire manipulation sequence compared to both VR‑only and leader–follower baselines. Specifically, baseline VR control suffered from erratic end‑effector trajectories when recovering from large Cartesian errors, and occasionally stalled near kinematic singularities. In cluttered scenes—most notably during the towel placement and clamp attachment subtasks—these instabilities led to frequent trajectory overshoot, oscillations, and safety‑margin violations. The leader–follower baseline, while more stable in free space, also suffered from reachability issues due to suboptimal joint configurations. These limitations contributed to frequent task failures under challenging spatial constraints. 

In contrast, our method maintained smooth and reliable performance across all subtasks. The integration of null-space guided configuration optimization helped mitigate abrupt joint updates and enabled more stable arm behavior, particularly during fine manipulation stages. Task success rates also improved: overall task completion rose to 60\% (24/40) in VR mode and 75\% (30/40) in leader–follower mode, compared to 45\% and 57\% for the respective baselines. 

To facilitate qualitative review, we have archived all experiment recordings in the supplementary materials, illustrating the smooth, coordinated behavior achieved by our framework under both teleoperation modalities.

\subsection{Failure Case Analysis and Discussion}
We analyzed task failures across all five subtasks to identify the limitations of our system. Errors during \textit{item unpacking} primarily occurred when objects within the cloth bag were occluded or improperly oriented, causing incorrect grasping. In the \textit{shelf organization} subtask, our null-space–augmented inverse kinematics (IK) method reliably positioned items into their designated slots. The infrequent failures that did occur typically resulted from transient tracking drift or calibration inaccuracies. For \textit{towel folding}, the main challenge emerged when lifting the folded towel off the table, as the layers tended to separate and slip, leading to partial unfolding. This instability directly affected the subsequent \textit{towel placement} step, as misaligned or unfolded edges frequently missed the rack bar, necessitating additional time for regrasping and realignment. The \textit{clamp attachment} subtask, despite demanding high precision, demonstrated robust performance due to the teleoperation system's accuracy, highlighting the effectiveness of our control approach. 

Overall, these results underscore the importance of robust IK solutions and coordinated control frameworks for precision manipulation tasks, while also pointing toward the necessity for enhanced strategies in handling deformable objects. Future efforts will focus on incorporating tactile feedback and adaptive grasp techniques to preserve structural integrity during manipulation, thereby improving real-world task success rates.

\section{CONCLUSIONS}
In this paper, we presented UniBiDex, a unified teleoperation framework that supports both VR and leader–follower inputs through a shared kinematic and safety-aware control module. Our system enables precise, real-time bimanual manipulation across diverse devices and tasks. User studies show that UniBiDex outperforms strong baselines in task success, motion smoothness, and robustness, particularly in contact-rich scenarios. We hope UniBiDex will lower the barrier to collecting large-scale, high-quality demonstration datasets and thereby accelerate progress in robot learning.

\bibliographystyle{ieeetr}
\bibliography{reference.bib}

\end{document}